\pdfoutput=1

\documentclass{article} %
\usepackage[preprint]{colm2025_conference}

\usepackage{microtype}
\usepackage{hyperref}
\usepackage{url}
\usepackage{booktabs}
\usepackage{lineno}
\usepackage{graphicx}
\usepackage{amsmath} 
\usepackage{makecell}
\usepackage{float}
\usepackage{multirow}
\usepackage{pgfplots}
\usepackage{pgfplotstable}
\usepackage{tikz}
\usepackage{longtable}
\usepackage{tabularx}
\usepackage{adjustbox}
\usepackage{array}  
\usepackage{makecell}
\usepgfplotslibrary{polar}
\usetikzlibrary{patterns,arrows.meta}
\pgfplotsset{compat=newest}

\definecolor{darkblue}{rgb}{0, 0, 0.5}
\hypersetup{colorlinks=true, citecolor=darkblue, linkcolor=darkblue, urlcolor=darkblue}

\newcommand{\methodname}{\textsc{BiasInspector}}
\newcommand{\benchmarkname}{\textsc{BiasBenchmark}}

\title{{\methodname}: Detecting Bias in Structured Data through LLM Agents}

\author{
Haoxuan Li$^{1}$~~~ 
Mingyu Derek Ma$^{2}$~~~ 
Jen-tse Huang$^{1}$~~~ 
Zhaotian Weng$^{1}$~~~ 
\textbf{Wei Wang}$^{2}$ \\
\medskip
\textbf{Jieyu Zhao}$^{1}$ \\
\medskip
$^1$University of Southern California~~~ 
$^2$University of California, Los Angeles \\
\texttt{\{lihaoxua, jh\_116, wengzhao, jieyuz\}@usc.edu}~~~ 
\texttt{\{ma, weiwang\}@cs.ucla.edu}
}

\begin{document}

\ifcolmsubmission
\linenumbers
\fi

\maketitle

\begin{abstract}
    Detecting biases in structured data is a complex and time-consuming task.
Existing automated techniques are limited in diversity of data types and heavily reliant on human case-by-case handling, resulting in a lack of generalizability.
Currently, large language model (LLM)-based agents have made significant progress in data science, but their ability to detect data biases is still insufficiently explored.
To address this gap, we introduce the first end-to-end, multi-agent synergy framework, {\methodname}, designed for automatic bias detection in structured data based on specific user requirements.
It first develops a multi-stage plan to analyze user-specified bias detection tasks and then implements it with a diverse and well-suited set of tools.
It delivers detailed results that include explanations and visualizations.
To address the lack of a standardized framework for evaluating the capability of LLM agents to detect biases in data, we further propose a comprehensive benchmark that includes multiple evaluation metrics and a large set of test cases.
Extensive experiments demonstrate that our framework achieves exceptional overall performance in structured data bias detection, setting a new milestone for fairer data applications.\footnote{Code and data are available at: \url{https://github.com/uscnlp-lime/BiasInspector}}%

\end{abstract}

\section{Introduction}
\label{sec:intro}

Data have been extensively utilized for various purposes, including model training, decision support, and personalized recommendations~\cite{ZHANG2018146}.
However, inevitable biases in data are the root causes of downstream biased behaviors in models~\cite{mehrabi2021survey}, including discriminatory decision-making~\cite{demartini2023data} and the perpetuation of social inequality~\cite{ferrara2023fairness}.
For example, biases in the MIMIC-IV dataset have been shown to result in predictive models unevenly distributing medical resources across different demographic and socioeconomic groups~\cite{kakadiaris2023evaluating}.
The detection of bias in structured data is defined as the process of analyzing and quantifying attribute distributions and correlations to identify unfairness or inaccuracies for specific subsets~\cite{shahbazi2023representation,balayn2021managing}.
It is an important step for improving model prediction reliability and fairness in fields such as healthcare \cite{li2024fairfml}, education \cite{chinta2024fairaied}, and finance \cite{zhou2024large}.

Biases in structured data vary in form and degree, making it hard to develop a universal automated detection method.
Despite advancements in automated bias detection, significant limitations remain.
Limited detection metrics struggle to identify certain types of bias or accurately measure their extent, making them difficult to apply to ineffective for application across diverse datasets~\cite{yuan2024fairerml,ahn2019fairsight}.
The requirement for specialized expertise forces users to understand bias detection concepts and possess programming skills, posing a significant barrier for non-expert users \cite{bellamy2018ai}.
Furthermore, the interpretability of detection results is insufficient, which primarily provide fixed numerical outputs while lacking intuitive visualization options, natural language descriptions of bias severity, and improvement suggestions~\cite{gujar2022genethos}.

Large Language Models (LLMs) have significantly enhanced the capabilities of autonomous agents to effectively address domain-specific tasks~\cite{xi2023risepotentiallargelanguage, Wang_2024}. Excelling in task planning~\cite{gu2024data} and workflow execution~\cite{sanger2023large}, LLMs are particularly well-suited for tackling data analysis tasks.
Recently, LLM-based agents have achieved remarkable progress in data analysis~\cite{majumder2024data,Hassan2023ChatGPTAY}.
However, due to the lack of flexible task planning for bias detection~\cite{yang2024matplotagent}, comprehensive bias detection tools~\cite{bordt2024data} and specific bias level descriptions~\cite{qiao2023taskweaver}, existing data science agents fail to meet the requirements for bias detection.
Thus, an automated agent capable of comprehensive bias detection in structured data is urgently needed to address these challenges.

In this work, we introduce {\methodname}, the first multi-agent synergy framework for detecting bias in structured data. As a fully end-to-end agent, it supports users in posing questions across all levels, ranging from generalized abstraction (e.g., \textit{Does insurance type influence the allocation of medical resources?}) to precise specifics (e.g., \textit{Using Cohen's d and Causal Effect to assess the strength of the correlation between \texttt{insurance\_type} and \texttt{icu\_stay}}). Our multi-agent collaborative framework executes iterative interactions across multiple stages, including data preprocessing, detection analysis, and visualization summarization. It continuously optimizes the formulation of plans and the application of tools, providing users with a comprehensive results report. The report not only includes professional metric results but also presents bias levels, visualizations, and dataset optimization recommendations in an easy-to-understand manner. It effectively addresses the needs of both technical experts and general users.

\begin{figure}[t]
    \centering
    \includegraphics[width=\textwidth]{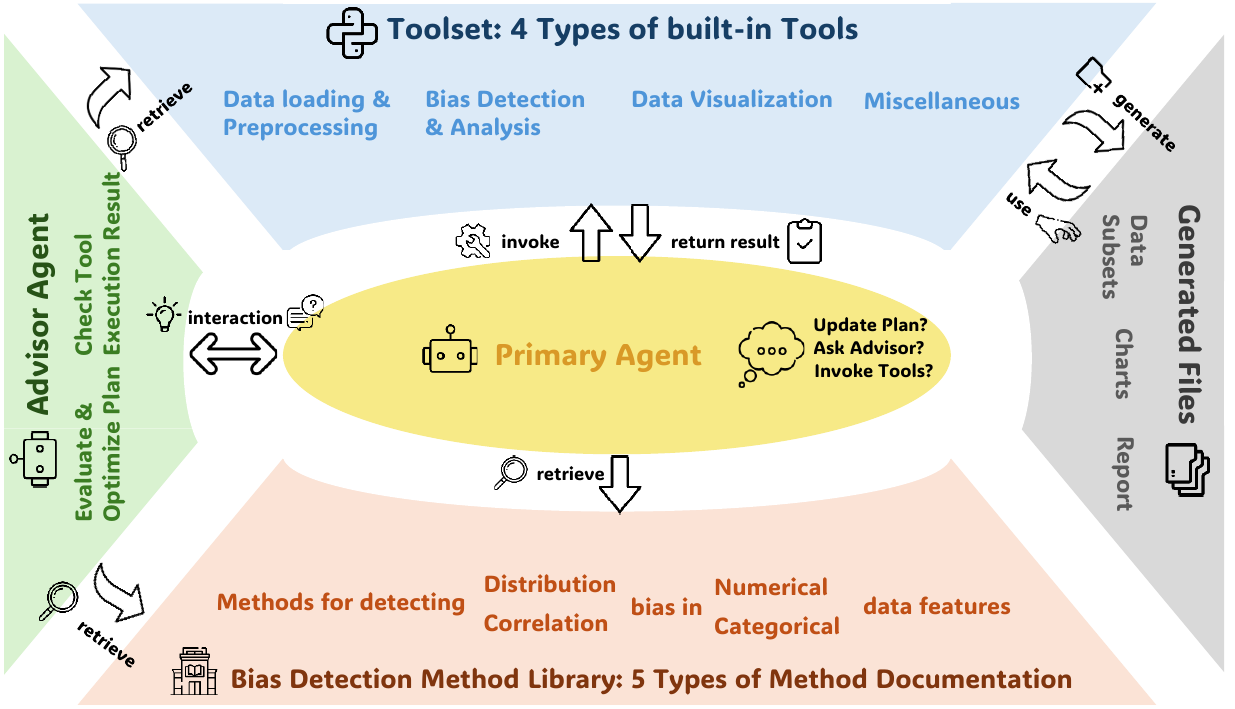} 
    \caption{Overview of the multi-agent architecture with a Primary and an Advisor Agent collaborating and invoking tools from the Toolset and Bias Detection Method Library.}
    \label{fig:system_overview}
\end{figure}

Faced with diverse data, the scarcity of detection metrics, the limited scalability of toolsets, and the complexity of operations are key factors hindering the development of previous technologies into general-purpose bias detection methods. To address these long-standing critical challenges in the field of bias detection, we retrieved metrics related to bias detection in structured data from the extensive literature. These metrics were refined into 46 predefined tools and 100 generatable tools, which were separately organized in the toolkit and method library, as shown in Fig.~\ref{fig:system_overview}. These two modules not only encompass a diverse range of detection tools, but also offer high extensibility. Data scientists and engineers can easily add new tools in the appropriate format to either module as needed. Additionally, the LLM's ability to directly invoke these tools eliminates the need for users to consult tool usage guides or possess programming skills.

Existing data analysis benchmarks ~\cite{majumder2024discoverybench} fail to evaluate the ability of LLM agents in detecting biases within structured data. Specifically, they lack metrics for assessing LLM agents' capability to detect multiple types of biases and their accuracy in measuring bias severity. To address this gap, we propose a new benchmark. We selected features with potential demographic biases from five commonly used datasets and designed a task set comprising 100 tasks for detecting various types of biases. Additionally, we developed an evaluation framework and designed an evaluation agent to separately assess performance from the end-result and intermediate-process perspectives.

We conducted extensive experiments using the newly proposed benchmark, including both End Result Evaluation and Intermediate Process Evaluation. The experimental results demonstrate that {\methodname} achieves an accuracy of up to 78\% in bias degree detection tasks and exhibits outstanding performance in six critical aspects such as planning and tooling. These findings underscore BiasInspector's significant contributions toward ensuring fairness in structured data applications.

In summary, our main contributions are:
\begin{enumerate}
    \item We introduce {\methodname}, the first LLM agent designed for detecting biases in structured data. It addresses the diverse needs of both professional and non-professional users in bias detection tasks.
    \item We proposed a benchmark that evaluates LLM agents' performance in detecting biases in structured data from both end-result and intermediate process perspectives.
    \item Extensive experiments demonstrate that {\methodname} achieves superior performance in structured data bias detection tasks, significantly outperforming existing agent baselines. Additionally, it consistently exhibits outstanding capabilities across multiple intermediate process dimensions.
\end{enumerate}

\section{Related Works}
\label{sec:related-works}

\paragraph{LLM Agents in Data Science}

Recently, some studies have leveraged LLM Agents in data science tasks, including data analysis and machine learning. In the field of data analysis, Majumder et al. ~\cite{majumder2024data} proposed using large generative models to automate iterative hypothesis generation, validation, and analysis for identifying data correlations. They further proposed a data-driven discovery benchmark~\cite{majumder2024discoverybench} to evaluate the capabilities of current LLMs in performing data discovery tasks. TaskWeaver ~\cite{qiao2023taskweaver} supports data analysis across diverse structures via code generation and plugins. HASSAN et al. ~\cite{Hassan2023ChatGPTAY} employed LLM Agents to facilitate user engagement with data. Bordt et al. ~\cite{bordt2024data} combined interpretable models with LLMs to generate detailed dataset summaries. MatPlotAgent ~\cite{yang2024matplotagent} focuses on automating data visualization using LLM Agents. In addition, some studies utilizing large language models to address machine learning tasks in data science also involve data cleaning and analysis~\cite{hong2024datainterpreterllmagent,guo2024dsagentautomateddatascience}.
Our work also involves correlation analysis and dataset visualization, but differs by providing comprehensive bias detection tools, autonomous planning, and detailed bias-level descriptions and recommendations.

\paragraph{Automated Bias Detection in Structured Data}

Recent studies have focused on developing automated methods to detect biases in structured datasets.
Prior works addressed bias quantification in synthetic data~\cite{gujar2022genethos}, examined data bias effects on model fairness~\cite{ahn2019fairsight}, and aimed to mitigate bias propagation in classifier training~\cite{kamiran2012data}. However, these studies rely on limited fixed metrics, restricting applicability across diverse datasets. FairerML~\cite{yuan2024fairerml}, an extensible platform integrating various fairness metrics, offers an interactive interface that allows users to select tools for bias analysis and visualization. AIF360~\cite{bellamy2018ai}, an open-source toolkit providing comprehensive bias measurement tools, requires extensive documentation navigation and advanced programming skills, thus posing barriers for non-technical users. In contrast, our method leverages an LLM Agent to automate the entire bias detection pipeline, offering richer and scalable toolsets, enhanced interpretability through natural language descriptions, and visualizations.

\section{{\methodname}}
\label{sec:agent}

\begin{figure}[t]
    \centering
    \includegraphics[width=\textwidth]{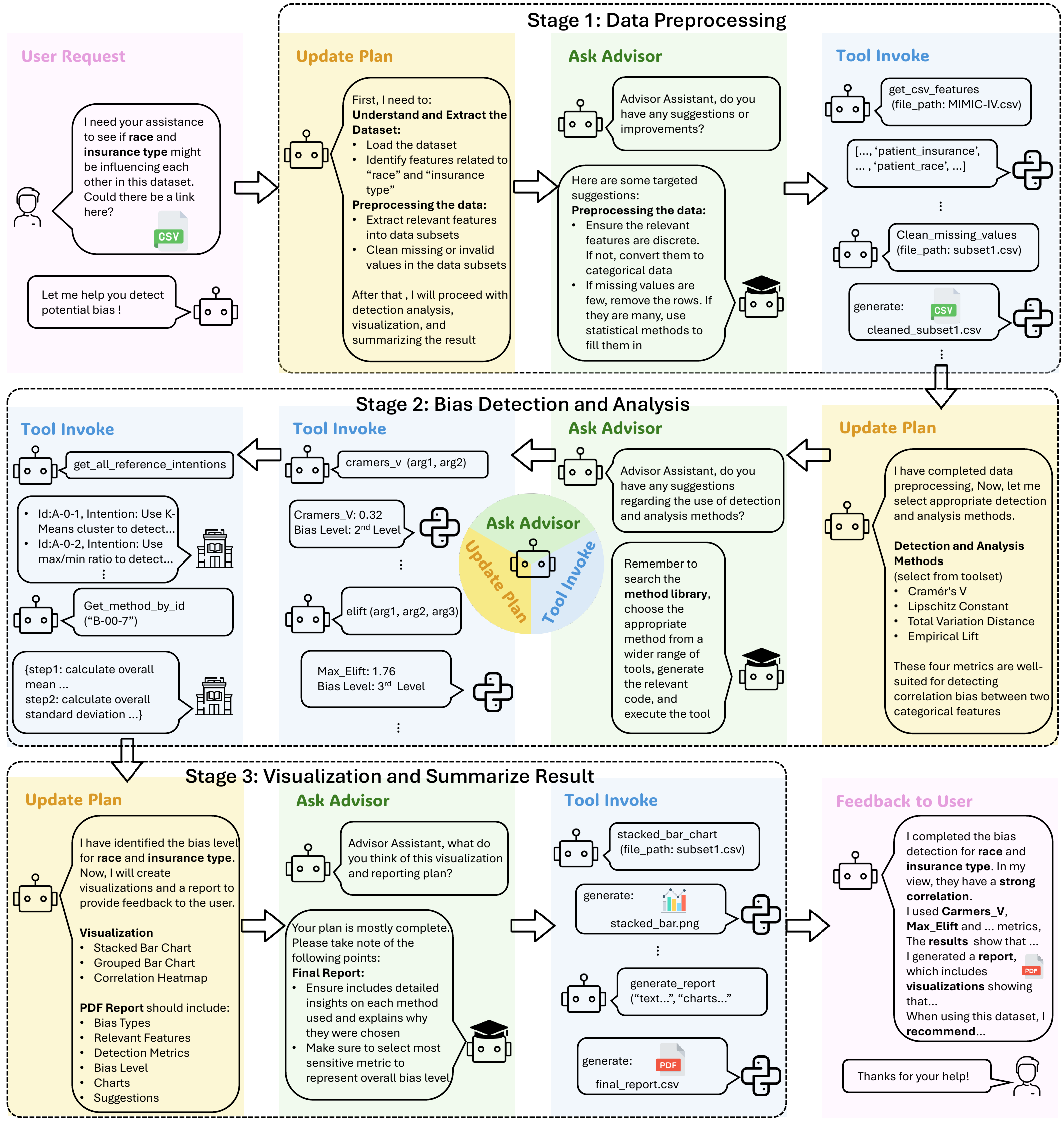} 
    \caption{Workflow overview: User Input, Data Preprocessing, Bias Detection and Analysis, Visualization and Summarization, and User Feedback. It is iterative rather than strictly sequential, allowing returns to previous stages based on user input or updated plans.}
    \label{fig:Flowchart}
\end{figure}

In this section, we introduce {\methodname}, a multi-agent system driven by LLM for detecting bias in structured data. 
We first discuss how bias manifests in structured data, then introduce the diverse set of tools supported by {\methodname}, its framework architecture, and detection process. 
This section illustrates how {\methodname} effectively achieves accurate and comprehensive bias detection in diverse bias representations.

\subsection{Collection of Bias Detection Methods and Construction of Toolset}
\paragraph{Rationale for Method Selection}

Bias in structured data includes distribution bias in individual features and correlation bias among multiple features, occurring in both numerical and categorical data. Numerical data exhibit varied distributions, while categorical data differ significantly in category counts. Such diversity prevents a single method from effectively addressing all bias forms, requiring a variety of specialized approaches tailored to specific bias manifestations.

Previous studies proposed numerous methods targeting specific data types or biases but lacked comprehensive coverage. Nevertheless, these methods offer valuable insights. Therefore, we systematically integrate relevant existing methods into a unified toolkit, enabling our Agent to comprehensively detect all forms of bias.

\paragraph{Method Compilation and Toolset Development}

To leverage previous research, we manually reviewed literature and employed large language models to extract structured data bias detection methods. Each method was decomposed into  explicit steps, stored uniformly in a JSON file. To facilitate accurate and efficient retrieval by the agent, we systematically annotated each method in the JSON file according to its corresponding bias type (distribution bias, correlation bias), data type (numerical, categorical), detection methodology, and application domain (e.g., medical, social sciences). This annotated resource, termed the Bias Detection Method Library, is detailed in Appendix~\ref{appendix:method-library}.

From this library, we compiled 100 methods covering all bias and data types, Further, we categorizing them into five representative scenarios, each corresponding to specific combinations of bias types and data types. For each scenario, we selected the five most representative methods, encapsulating them as callable Python functions. These constitute our Toolset, comprising 25 distinct bias detection tools.

\paragraph{RAG-Based Tool Retrieval and Execution}

Upon receiving user input, our Agent system automatically determines the bias type and data type scenario relevant to the current task, based on the user's instructions and the results of data loading and preprocessing. Using Retrieval-Augmented Generation (RAG), the Agent selects suitable bias detection tools in two steps: first retrieving applicable methods from the Bias Detection Method Library, then directly invokes predefined Python functions for methods available in the Toolset; for those methods not predefined in the Toolset, the Agent automatically generates executable code to perform the detection.

The Bias Detection Method Library is designed for extensibility, enabling users to easily integrate new methods using large language models through a consistent process. This ensures the Agent continuously benefit from increasingly comprehensive and advanced bias detection techniques in future applications.

Compared to traditional manual approaches, where domain experts typically select limited tools—potentially lacking sensitivity to certain biases—and must manually write or invoke detection code, our Agent system swiftly and automatically identifies multiple suitable tools from an extensive library, significantly enhancing comprehensiveness. Additionally, automating tool invocation and code generation markedly improves detection efficiency. Thus, the automated and expandable toolset enables our Agent to fully leverage its advantages in structured data bias detection.

\subsection{Agent System Architecture}
\paragraph{Functional Toolset}

To construct a comprehensive agent system for structured data bias detection, we developed a complete suite of functionalities, including data preprocessing, bias analysis, result visualization, and reporting. Since practical tasks require dataset-specific strategies, we implemented numerous functional tools suitable for various scenarios. These tools are encapsulated as predefined functions in a unified Toolset for direct Agent invocation. Currently, the Toolset contains 46 distinct tools, detailed in Appendix~\ref{sec:appendix-tools}.

\paragraph{Multi-agent Interaction}

Bias detection is typically an iterative reasoning-action process, where the Agent continuously updates execution plans based on user interactions and tool feedback. Detection tasks involve multiple stages, including data preprocessing, bias analysis, and result visualization, each potentially requiring single or combined tool usage. Such complexity imposes significant demands on large language models for planning, outcome interpretation, feedback, and decision-making.

To address potential limitations of large language models and enhance task completion rate and quality, we designed a multi-agent framework comprising a Primary Agent and an Advisor Agent. The Primary Agent directly interacts with users, formulates execution plans, invokes necessary tools, and consults the Advisor Agent at key stages. The Advisor Agent reviews and refines these plans, identifies oversights, recommends tool invocations, analyzes results, and provides strategies for anomalies or errors. Detailed roles of both agents are provided in Appendix~\ref{sec:appendix-agent-info}.

\subsection{Agent Execution Workflow}
This section describes the operational workflow for bias detection tasks, as illustrated in Figure ~\ref{fig:Flowchart}, comprising five key steps: user input, data preprocessing, bias detection and analysis, result visualization and summarization, and user feedback.

\paragraph{Step 1: User Input}

Users provide a structured dataset and bias-related questions or task instructions. {\methodname} supports varying expertise levels, accommodating general bias inquiries or explicit tool-specific instructions.

\paragraph{Step 2: Data Preprocessing}

Upon receiving data, the Agent formulates and executes a preprocessing plan, including reading data, identifying features and types, extracting relevant columns, handling missing values and outliers, and constructing a processed subset for downstream analysis.

\paragraph{Step 3: Bias Detection and Analysis}

The Agent then develops and executes a bias detection plan, selecting suitable methods from the Toolset and Bias Detection Method Library, sequentially executing them, and obtaining numerical metrics indicating bias severity.

\paragraph{Step 4: Result Visualization and Summarization}

The Agent formulates and executes a visualization strategy, selecting suitable representations to clearly illustrate results. It analyzes metrics, summarizes bias severity, and formulates targeted recommendations. All relevant information—including bias types, associated features, detection tools and metrics, bias severity levels, visualizations, and recommendations—is consolidated into a comprehensive PDF detection report.

\paragraph{Step 5: Result Feedback to Users}

Finally, the Agent presents detection results through natural language explanations and the PDF report, proactively engaging users for additional comments or requirements, ensuring alignment with user expectations.

\section{{\benchmarkname}}
\label{sec:benchmark}

Effective evaluation of agent capability in structured data bias detection remains underexplored. Existing data analysis benchmarks~\cite{majumder2024discoverybench} inadequately assess diverse bias scenarios and lack accuracy evaluation for bias severity estimation. To address these limitations, we propose a novel benchmark framework. In this section, We first describe benchmark construction, then discuss two evaluation dimensions: end-result evaluation and intermediate-process evaluation.

\subsection{Benchmark Setup and Task Design}
\paragraph{Dataset Selection}

We selected five structured datasets widely used in prior bias mitigation research~\cite{Hort2022BiasMF}: Adult~\cite{adult_2}, COMPAS~\cite{doi:https://doi.org/10.1002/9781119184256.ch3}, Statlog~\cite{statlog_(german_credit_data)_144}, MIMIC-IV~\cite{Johnson2023}, and Student Performance~\cite{student_performance_320}. These datasets cover diverse domains (socioeconomic, criminal justice, finance, healthcare, education), vary from hundreds to tens of thousands of instances, and include 10 to 33 categorical and numerical features. Thus, they are highly suitable for evaluating an agent’s bias detection capability in structured data. Dataset details are provided in Appendix~\ref{sec:appendix-datasets}.

\paragraph{Task Set Construction}

We selected 100 demographic-related features or feature combinations from the datasets, covering categorical and numerical data. Individual features were classified as potentially exhibiting distribution bias, while feature combinations were classified as involving correlation bias. Using a large language model, we created diverse bias detection queries for each case, varying linguistic expressions to simulate human questioning styles. Queries include explicitly stated bias types (distribution or correlation) and intentionally ambiguous forms, termed implication bias, reflecting real-world user scenarios. Task set details and examples are provided in Appendix~\ref{sec:appendix-tasksets}.

\paragraph{Bias Severity Quantification}

Bias detection in structured datasets involves not only identifying bias presence but also accurately quantifying its severity,which is typically challenging. To enable quantitative severity assessment, we defined five bias levels, ranging from 'most balanced' to 'most biased.'

We categorized bias manifestations into five representative scenarios based on combinations of data types (categorical, numerical) and bias types (distribution, correlation). For each scenario, we selected five widely-used bias detection metrics, developed corresponding detection scripts, and mapped metric values to the predefined bias severity levels.

To ensure mapping accuracy, we manually constructed synthetic datasets exhibiting varying degrees of bias, data scales, and feature counts. Through iterative testing and refinement, we verified that our scripts precisely map different bias severities to correct levels. These verified mappings serve as the ground truth for outcome evaluation. %

\subsection{End Result Evaluation}

We developed an automated evaluation framework to assess the agent’s bias detection performance from an outcome perspective. For each task, the framework identifies the involved feature or feature combination, maps it to the corresponding bias scenario, and applies five scenario-specific detection tools, each producing a predicted bias level. We select the maximum predicted level as the reference value, assuming it represents the most sensitive bias detection.

We then compare this reference with the agent's predicted bias level. The absolute difference quantitatively measures the agent's outcome-level detection accuracy.

\subsection{Intermediate Process Evaluation}
When evaluating an agent's bias detection capability, it is essential to consider not only final outcomes but also the reasoning, planning, and decision-making processes throughout task execution. To achieve a comprehensive assessment, we further evaluate the agent from a process-oriented perspective, focusing on five aspects: user communication, task planning, tool invocation, dynamic plan adjustment, and result analysis.

We developed an agent-based automated evaluation system, defining five performance rating levels (Excellent, Proficient, Adequate, Mediocre, Unsatisfactory) with explicit criteria across these dimensions. This system automatically analyzes the agent’s operational logs and generates a PDF report detailing ratings and supporting rationale for each evaluated task. Further details are provided in Appendix~\ref{sec:appendix-agent-eval}.

To validate the automated evaluation's reliability, we recruited human evaluators to independently review execution logs and assess the agent's performance along the same five dimensions. Correlation analysis comparing automated ratings with human assessments confirmed the effectiveness and reliability of our evaluation method. Additional details are provided in Appendix~\ref{sec:appendix-human-eval}.

\section{Experiments}
\label{sec:experiments}

\subsection{Experimental Setup}
To systematically evaluate {\methodname}, we designed several framework-model combinations and conducted experiments using the benchmark from Section~4.

\paragraph{Framework Settings}  
We evaluated four agent frameworks: (1) the full {\methodname} (Primary and Advisor Agents); (2) single-agent {\methodname} (Primary Agent only); (3) ReAct-based agent~\cite{yao2023react} equipped with the same toolset as BiasInspector; and (4) Self-reflection agent, relying solely on model reasoning with tool access limited to basic data reading.

\paragraph{Model Choices}  
We conducted experiments with two widely-used large language models: GPT-4o and Llama 3.3 70B. Decoding temperature was set to 0 for output stability.

\paragraph{Evaluation Metrics}  
Each framework-model combination is evaluated from two perspectives:

\begin{itemize}
    \item \textit{End-result  evaluation.} To measure accuracy, we compute the Mean Absolute Error (MAE) between agent-predicted and ground-truth bias levels across tasks. For intuitive interpretation, we convert MAE into an \textit{Average Similarity Score} $S_{\text{avg}}$, defined as $S_{\text{avg}} = \frac{1}{n}\sum_{i=1}^{n}\left(1 - \frac{|x_i - y_i|}{4}\right)\times 100\%$, where $x_i$ and $y_i$ denote the predicted and ground-truth bias levels for task $i$. Higher scores indicate greater accuracy.

    \item \textit{Intermediate-process evaluation.} Using our automated GPT o3-mini-based evaluation system, we assess each task across five dimensions: user interaction, planning, tool invocation, dynamic adaptation, and result summarization. We report the average score over all tasks, reflecting overall workflow execution capability.
\end{itemize}

\subsection{End Result Performance}
\begin{table}[H]
\renewcommand{\arraystretch}{1.1}
\begin{center}
\small
\resizebox{\linewidth}{!}{%
\begin{tabular}{cccccc}
\toprule
\makecell{\textbf{Model}} 
& \makecell{\textbf{Bias Type}} 
& \makecell{\textbf{BiasInspector} \\ \textbf{(Multi-Agent)}} 
& \makecell{\textbf{BiasInspector} \\ \textbf{(Single-Agent)}} 
& \makecell{\textbf{ReAct-Based} \\ \textbf{Agent}} 
& \makecell{\textbf{Self-Reflection} \\ \textbf{Agent}} \\
\midrule

\multirow{4}{*}{\raisebox{-0.4ex}{GPT-4o}}
& Overall       & \textbf{77.53} & \underline{76.52} & 67.68 & 60.61 \\
& Distribution  & \underline{84.46} & \textbf{86.49} & 72.97 & 60.14 \\
& Correlation   & \textbf{72.70} & \underline{64.80} & 60.86 & 59.54 \\
& Implication   & \underline{76.04}  & \textbf{81.25}  & 72.92  & 65.62  \\[2pt]

\multirow{4}{*}{\raisebox{-0.4ex}{LLaMA 3.3 70B}} 
& Overall       & \textbf{71.97} & \underline{71.46} & 63.13 & 66.67 \\
& Distribution  & \underline{74.32} & 72.30 & \textbf{77.03} & 66.89 \\
& Correlation   & \textbf{71.79} & \underline{70.51} & 55.13 & 63.46 \\
& Implication   & 68.75   & \underline{70.83}   & 55.21 & \textbf{71.88}   \\

\bottomrule
\end{tabular}%
}
\end{center}
\caption{End-results performance across different agent configurations and LLMs. All values represent average similarity scores (\%).}
\label{tab:overall_end_results}
\end{table}

In Table~\ref{tab:overall_end_results}, we present Average Similarity Scores for model-framework combinations evaluated on the full task set and subsets (distribution bias, correlation bias, implication bias). This metric effectively reflects bias detection accuracy for each combination.

\paragraph{Comparative Analysis of Agent Architectures} 

Our analysis shows BiasInspector consistently achieves 70\%-80\% similarity scores, significantly outperforming ReAct and Self-Reflection agents. The ReAct Agent shows instability with notable accuracy fluctuations across bias types. While Multi-Agent and Single-Agent BiasInspector perform similarly overall, the Multi-Agent variant achieves 8 points higher accuracy on complex correlation bias tasks with GPT-4o, highlighting collaborative agent benefits.

\paragraph{Comparative Analysis of LLMs} 

The GPT series consistently outperforms Llama 3.3 70B when powering BiasInspector. Notably, GPT-4o achieves over 10 percentage points higher accuracy than Llama 3.3 70B on distribution bias tasks in both Multi-Agent and Single-Agent architectures. This underscores that powerful LLMs like GPT-4o better leverage BiasInspector’s planning and tooling capabilities, substantially enhancing structured data bias detection effectiveness.

\subsection{Intermediate Process Performance}

\definecolor{colorA}{RGB}{70,130,180}
\definecolor{colorB}{RGB}{102,170,100}
\definecolor{colorC}{RGB}{150,110,50}
\definecolor{colorD}{RGB}{80,80,150}

\begin{figure}[H]
\centering

\begin{tikzpicture}
\node (legend) at (0,0) {
    \begin{tikzpicture}
    \begin{axis}[
        hide axis,
        xmin=0, xmax=100,
        ymin=0, ymax=100,
        legend columns=4,
        legend style={draw=none, font=\scriptsize, column sep=5pt},
    ]
    \addlegendimage{colorA, mark=*, mark options={scale=0.5}, fill=colorA, fill opacity=0.1}
    \addlegendentry{BiasInspector (Multi)}
    \addlegendimage{colorB, mark=*, mark options={scale=0.5}, fill=colorB, fill opacity=0.1}
    \addlegendentry{BiasInspector (Single)}
    \addlegendimage{colorC, mark=*, mark options={scale=0.5}, fill=colorC, fill opacity=0.1}
    \addlegendentry{ReAct Agent}
    \addlegendimage{colorD, mark=*, mark options={scale=0.5}, fill=colorD, fill opacity=0.1}
    \addlegendentry{Self-Reflection Agent}
    \end{axis}
    \end{tikzpicture}
};
\end{tikzpicture}
\vspace{-0.6em}

\begin{minipage}{0.45\textwidth}
\centering
\begin{tikzpicture}
\begin{polaraxis}[
    width=5.0cm,
    xtick distance=60,
    xticklabel style={font=\scriptsize},
    axis line style={draw=none},
    xticklabels={Summarization, Integration, Communication, Planning, Tooling, Adaptivity},
    ymin=0, ymax=100,
    grid=both,
    clip=false,
    yticklabels=\empty,
    every axis plot/.append style={mark=*,mark options={scale=0.5}},
    extra description/.code={
        \node[font=\scriptsize, anchor=south] at (axis direction cs:300,140) {Summarization};
        \foreach \val in {20,40,60,80,100}{
            \node[font=\scriptsize, anchor=south west] at (axis direction cs:22.5,{0.8*\val}) {\val};
        }
    }
]
\addplot+[color=colorA,fill=colorA,fill opacity=0.1] coordinates {
    (0,90) (60,90) (120,92) (180,88) (240,89) (300,91)
};

\addplot+[color=colorB,fill=colorB,fill opacity=0.1] coordinates {
    (0,88) (60,87) (120,90) (180,87) (240,84) (300,90)
};

\addplot+[color=colorC,fill=colorC,fill opacity=0.1] coordinates {
    (0,80) (60,73) (120,85) (180,79) (240,78) (300,84)
};

\addplot+[color=colorD,fill=colorD,fill opacity=0.1] coordinates {
    (0,53) (60,62) (120,63) (180,43) (240,43) (300,65)
};
\end{polaraxis}
\end{tikzpicture}
\end{minipage}
\hfill
\begin{minipage}{0.45\textwidth}
\centering
\begin{tikzpicture}
\begin{polaraxis}[
    width=5.0cm,
    xtick distance=60,
    xticklabel style={font=\scriptsize},
    axis line style={draw=none},
    xticklabels={Summarization, Integration, Communication, Planning, Tooling, Adaptivity},
    ymin=0, ymax=100,
    grid=both,
    clip=false,
    yticklabels=\empty,
    every axis plot/.append style={mark=*,mark options={scale=0.5}},
    extra description/.code={
        \node[font=\scriptsize, anchor=south] at (axis direction cs:300,140) {Summarization};
        \foreach \val in {20,40,60,80,100}{
            \node[font=\scriptsize, anchor=south west] at (axis direction cs:22.5,{0.8*\val}) {\val};
        }
    }
]
\addplot+[color=colorA,fill=colorA,fill opacity=0.1] coordinates {
    (0,80) (60,82) (120,83) (180,79) (240,76) (300,78)
};

\addplot+[color=colorB,fill=colorB,fill opacity=0.1] coordinates {
    (0,75) (60,75) (120,80) (180,73) (240,69) (300,76)
};

\addplot+[color=colorC,fill=colorC,fill opacity=0.1] coordinates {
    (0,75) (60,73) (120,78) (180,72) (240,74) (300,76)
};

\addplot+[color=colorD,fill=colorD,fill opacity=0.1] coordinates {
    (0,43) (60,53) (120,49) (180,31) (240,30) (300,55)
};
\end{polaraxis}
\end{tikzpicture}
\end{minipage}

\vspace{0.5em}
\begin{minipage}{0.50\textwidth}
\centering
\scriptsize \textbf{(a)} Execution performance under GPT-4o
\end{minipage}
\hfill
\begin{minipage}{0.45\textwidth}
\centering
\scriptsize \textbf{(b)} Execution performance under LLaMA 3.3 70B
\end{minipage}

\caption{Intermediate process performance of four agent frameworks.}
\label{fig:dual_radar}
\end{figure}
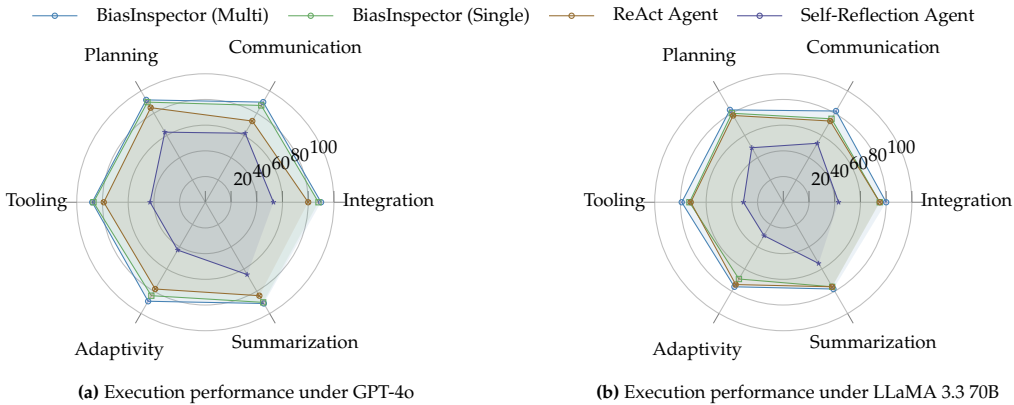

Figure~\ref{fig:dual_radar} presents Intermediate Process Performance of all model-agent architecture combinations across six dimensions: Integration, Communication, Planning, Tooling, Adaptivity, and Summarization, clearly illustrating BiasInspector’s superior performance.

\paragraph{Analysis of Performance Across Individual Dimensions} BiasInspector notably excels in Communication, Planning, and Summarization, achieving scores over 90 (GPT-4o) and over 80 (Llama 3.3 70B). It also demonstrates strong Tooling and Adaptivity, with minor gaps (1–2 points) from top scores, confirming its consistent excellence across dimensions.

\paragraph{Comparative Analysis of Different Agent Architectures} BiasInspector consistently surpasses ReAct agents by 5–15 points in key dimensions like Planning and Tooling, and significantly outperforms Self-Reflection agents. This highlights BiasInspector’s superior design and effective utilization of large language models, explaining its higher practical bias detection accuracy. Additionally, the Multi-Agent BiasInspector slightly outperforms the Single-Agent variant across dimensions, validating the benefits of multi-agent collaboration.

\paragraph{Performance Comparison Across Different LLMs} Figure~\ref{fig:dual_radar}(a) shows GPT-4o-powered agents outperform Llama 3.3 70B agents (Figure~\ref{fig:dual_radar}(b)) by about 10 points in all dimensions. This confirms advanced LLMs like GPT-4o significantly boost agent performance, effectively leveraged by BiasInspector.

\section{Conclusion}
\label{sec:conclusion}

In this work, we introduce {\methodname}, an LLM-driven multi-agent framework for automated bias detection in structured data. Collaborative agents formulate multi-stage task plans based on user requirements, using diverse analytical tools to deliver detailed bias detection results. To address the lack of standardized evaluation for LLM-based bias detection, we propose a comprehensive benchmark with multiple metrics and extensive test cases. Experiments shows {\methodname} achieves exceptional bias detection performance, ensuring robust data fairness and setting a benchmark for future research.

\bibliography{colm2025_conference}
\bibliographystyle{colm2025_conference}

\appendix
\section{Bias Detection Method Library}
\label{appendix:method-library}

To support flexible and reproducible bias detection, we developed a structured library of detection methods encoded in \texttt{JSON} format. Each method entry includes metadata and a set of structured procedural steps. Specifically, each record contains:

\begin{itemize}
    \item \textbf{id}: A unique identifier for the method entry.
    \item \textbf{intention}: A high-level description of the goal, specifying the data type, analytical approach, and bias context.
    \item \textbf{method}: A dictionary of ordered steps detailing the procedure for detecting bias using tools or algorithms.
    \item \textbf{title}: The title of the reference publication.
    \item \textbf{article link}: A URL linking to the original article.
    \item \textbf{field}: The academic or application domain.
    \item \textbf{year}: The year of publication.
\end{itemize}

Table~\ref{tab:method-library-sample} presents a sample excerpt from this library.
\begin{table}[H]
\centering
\caption{Excerpt from the JSON-based bias detection method library.}
\label{tab:method-library-sample}
\renewcommand{\arraystretch}{1.15}
\begin{tabular}{|p{1.8cm}|p{10.5cm}|}
\hline
\textbf{ID} & \textbf{Content Summary} \\
\hline

\texttt{A-0-1} &
\textbf{Intention:} Detect distribution bias in a categorical feature (gender) within the healthcare domain using K-Means clustering and an entropy-based balance measure. \newline
\textbf{Steps:}
\begin{itemize}
    \item Apply K-Means clustering to gender and height attributes (10 clusters).
    \item Use MapReduce to process clusters and extract gender category distributions.
    \item Calculate Shannon entropy of gender: $H = -\sum (c_i/n) \log(c_i/n)$.
    \item Compute bias balance score: $Balance = H / \log k$.
\end{itemize}
\textbf{Title:} \textit{Automated Bias Detection within the Cardiovascular Disease Dataset using MapReduce Framework with Balance Measure} \newline
\textbf{Field:} Healthcare \newline
\textbf{Year:} 2024 \newline
\textbf{Link:} \url{https://ijisae.org/index.php/IJISAE/article/view/5688/4429} \\
\hline

\texttt{A-0-2} &
\textbf{Intention:} Detect distribution bias in categorical features using the max/min ratio of category frequencies in text classification datasets. \newline
\textbf{Steps:}
\begin{itemize}
    \item Detect categorical (non-numeric) features.
    \item Compute relative frequency for each category.
    \item Identify max and min category ratios; extreme values indicate bias.
    \item Calculate the max/min ratio as the bias score.
    \item Assign bias level using thresholds: $>100$ (Extreme), $>10$ (Significant), etc.
\end{itemize}
\textbf{Title:} \textit{Selection of the Most Relevant Terms Based on a Max-Min Ratio Metric for Text Classification} \newline
\textbf{Field:} Text Classification \newline
\textbf{Year:} 2018 \newline
\textbf{Link:} \url{https://www.sciencedirect.com/science/article/abs/pii/S0957417418304457} \\
\hline

\end{tabular}
\end{table}

\section{Information on Tools in the Toolkit}
\label{sec:appendix-tools}

We have created four categories of tools, totaling 46 tools in the toolkit, including:

\begin{itemize}
    \item 7 commonly used data loading and preprocessing tools, designed to meet various preprocessing needs for structured datasets.
    \item 25 bias detection tools, with 5 commonly used metric detection tools provided for each type of bias manifestation.
    \item 9 data visualization tools, offering comprehensive chart representations for different data types.
    \item 5 miscellaneous tools, designed for user interaction, reading the reference methods library, executing dynamically generated tool code, and generating PDF detection reports.
\end{itemize}

Table~\ref{tab:tools_info_in_toolset} presents each tool and its functional description.
\begin{longtable}{|p{5cm}|p{8cm}|}
\caption{Description of tools in toolset.} \\
\hline
\multicolumn{2}{|c|}{\rule{0pt}{3ex}\textbf{Data Loading and Preprocessing Tools}\rule{0pt}{3ex}} \\
\hline
\textbf{Tool} & \textbf{Description} \\
\hline
\endfirsthead
\multicolumn{2}{c}{\textit{(Continued from previous page)}} \\
\hline
\textbf{Tool} & \textbf{Description} \\
\hline
\endhead
\hline \multicolumn{2}{|r|}{\textit{Continued on next page}} \\
\hline
\endfoot
\hline
\endlastfoot
get\_csv\_features & Reads a CSV file and returns all feature names (column names) \\
\hline
load\_csv\_file & Loads a CSV file and returns it as a Pandas DataFrame \\
\hline
extract\_single\_column & Extracts a single column from a CSV file and saves it as a new dataset \\
\hline
extract\_two\_columns & Extracts two columns from a CSV file and saves them as a new dataset \\
\hline
clean\_missing\_values & Cleans missing or invalid values from specified columns and saves the cleaned dataset \\
\hline
normalize\_or\_standardize\_data & Applies 'normalize' or 'standardize' to a specified column and saves the result as a new dataset \\
\hline
group\_and\_aggregate & Groups the data by a specified column and applies an aggregation function on another column \\
\hline
\multicolumn{2}{|c|}{\rule{0pt}{3ex}\textbf{Detect and Analysis Tools}\rule{0pt}{3ex}} \\
\hline
\textbf{Tool} & \textbf{Description} \\
\hline
categorical\_distribution\_shannon \_balance & Analyzes the distribution bias of a categorical column using Shannon entropy and Balance metric \\
\hline
categorical\_distribution\_shannon \_balance & Analyzes the distribution bias of a categorical column using Shannon entropy and Balance metric. \\
\hline
categorical\_distribution\_max \_min\_ratio & Analyzes the distribution bias of a categorical column using the max/min ratio of categories. \\
\hline
categorical\_distribution\_entropy & Analyzes the distribution bias of a categorical column using Shannon entropy and normalized entropy. \\
\hline
categorical\_distribution\_gini & Analyzes the distribution bias of a categorical column using the Gini Index with Laplace smoothing and sample size correction. \\
\hline
categorical\_distribution\_relative \_risk & Analyzes the distribution bias of a categorical column using relative risk by comparing observed and expected frequencies. \\
\hline
numerical\_distribution\_skewness & Analyzes the distribution bias of a numerical column using skewness to assess asymmetry in the data. \\
\hline
numerical\_distribution\_kurtosis & Analyzes the distribution bias of a numerical column using kurtosis to assess the "tailedness" of the distribution. \\
\hline
numerical\_distribution\_outlier & Analyzes the distribution bias of a numerical column using Z-score outlier detection to assess the percentage of outliers. \\
\hline
numerical\_distribution\_cohens \_d\_mad & Analyzes the distribution bias of a numerical column using Cohen's d, calculated with Median Absolute Deviation (MAD) for robustness against outliers. \\
\hline
numerical\_distribution\_quantile \_deviation & Analyzes the distribution bias of a numerical column using quantile deviation, calculated as the ratio of Q3-Q2 to the interquartile range (IQR). \\
\hline
categorical\_categorical \_correlation\_cramers\_v & Analyzes the correlation bias between two categorical columns using Cramér's V, which measures association strength based on the chi-square statistic. \\
\hline
categorical\_categorical \_correlation\_elift & Analyzes the correlation bias between two categorical columns using Elift, calculated as the ratio of confidence(X -> Y) to confidence(Y). \\
\hline
categorical\_categorical \_correlation\_statistical\_parity & Analyzes the correlation bias between two categorical columns using statistical parity and Z-scores to assess differences in proportions between groups. \\
\hline
categorical\_categorical \_correlation\_lipschitz & Analyzes the correlation bias between two categorical columns using a Lipschitz function to measure distribution loss differences between groups. \\
\hline
categorical\_categorical \_correlation\_total\_variation & Analyzes the correlation bias between two categorical columns using Total Variation Distance (TVD) to measure the difference between group-specific and overall distributions. \\
\hline
categorical\_numerical\_correlation \_max\_abs\_mean & Analyzes the correlation bias between a categorical and a numerical column by computing the standardized mean for each category and evaluating the maximum absolute mean (N value). \\
\hline
categorical\_numerical\_correlation \_cohens\_d & Analyzes the correlation bias between a categorical and a numerical column using Cohen's d to measure the effect size based on an independent t-test. \\
\hline
categorical\_numerical\_correlation \_standardized\_difference & Analyzes the correlation bias between a categorical and a numerical column using Standardized Difference (SD), calculated with mean and Median Absolute Deviation (MAD). \\
\hline
categorical\_numerical\_correlatio n\_causal\_effect & Analyzes the correlation bias between a categorical treatment and a numerical outcome using causal inference to estimate the Average Causal Effect (ACE). \\
\hline
categorical\_numerical\_correlation \_pse & Analyzes the correlation bias between a categorical and a numerical column using Path-Specific Effect (PSE), along with the Average Direct Effect (ADE) and Average Indirect Effect (AIE). \\
\hline
numerical\_numerical\_correlation \_pearson & Analyzes the correlation bias between two numerical columns using Pearson correlation to measure the strength of their linear relationship. \\
\hline
numerical\_numerical\_correlation \_nmi & Analyzes the correlation bias between two numerical columns using Normalized Mutual Information (NMI) after discretizing the data into bins. \\
\hline
numerical\_numerical\_correlation \_hgr\_approximation & Analyzes the correlation bias between two numerical columns using HGR approximation and chi-squared 
 divergence, combining Pearson correlation and Kernel Density Estimation (KDE) to assess the strength of the relationship. \\

\hline
numerical\_numerical\_correlation \_wasserstein & Analyzes the correlation bias between two numerical columns using Wasserstein-2 distance to measure the differences between their distributions. \\
\hline
numerical\_numerical\_correlation \_hsic & Analyzes the correlation bias between two numerical columns using the Hilbert-Schmidt Independence Criterion (HSIC) with Radial Basis Function (RBF) kernels to measure their dependence. \\
\hline
\multicolumn{2}{|c|}{\rule{0pt}{3ex}\textbf{Visualization Tools}\rule{0pt}{3ex}} \\
\hline
\textbf{Tool} & \textbf{Description} \\
\hline
plot\_bar\_chart & Generates a bar chart for a specified column in a CSV file and saves it as an image. \\
\hline
plot\_pie\_chart & Generates a pie chart for a specified column in a CSV file and saves it as an image. \\
\hline
plot\_horizontal\_bar\_chart & Generates a horizontal bar chart for a specified column in a CSV file and saves it as an image. \\
\hline
plot\_treemap & Generates a treemap for a specified column in a CSV file and saves it as an image. \\
\hline
plot\_heatmap & Generates a heatmap for the frequency distribution of a specified column in a CSV file and saves it as an image. \\
\hline
plot\_correlation\_heatmap & Generates a correlation heatmap for multiple specified columns in a CSV file and saves it as an image. \\
\hline
plot\_stacked\_bar\_chart & Generates a stacked bar chart for two specified categorical columns in a CSV file and saves it as an image. \\
\hline
plot\_grouped\_bar\_chart & Generates a grouped bar chart for two specified categorical columns in a CSV file and saves it as an image. \\
\hline
plot\_box\_plot & Generates a box plot for a categorical column and a numeric column in a CSV file and saves it as an image. \\
\hline
\multicolumn{2}{|c|}{\rule{0pt}{3ex}\textbf{Miscellaneous Tools}\rule{0pt}{3ex}} \\
\hline
\textbf{Tool} & \textbf{Description} \\
\hline
get\_user\_input\_tool & Captures user input dynamically during an interaction and formats it as a dictionary to be added to the Agent's conversation. \\
\hline
get\_all\_reference\_intentions & Retrieves all intentions from the references JSON file, including each reference's id and corresponding intention. \\
\hline
get\_reference\_method\_by\_id & Retrieves the method for a specific reference by ID from the references JSON file. \\
\hline
generate\_bias\_report\_pdf & Generates a bias detection report in PDF format, combining text and charts as specified in the input. \\
\hline
execute\_python\_code & Executes the provided Python code and returns the printed output, if any. 

\label{tab:tools_info_in_toolset}
\end{longtable}

\section{Instructions for the Two Agents}
\label{sec:appendix-agent-info}

\begin{figure}[H]
    \centering
    \includegraphics[width=\textwidth]{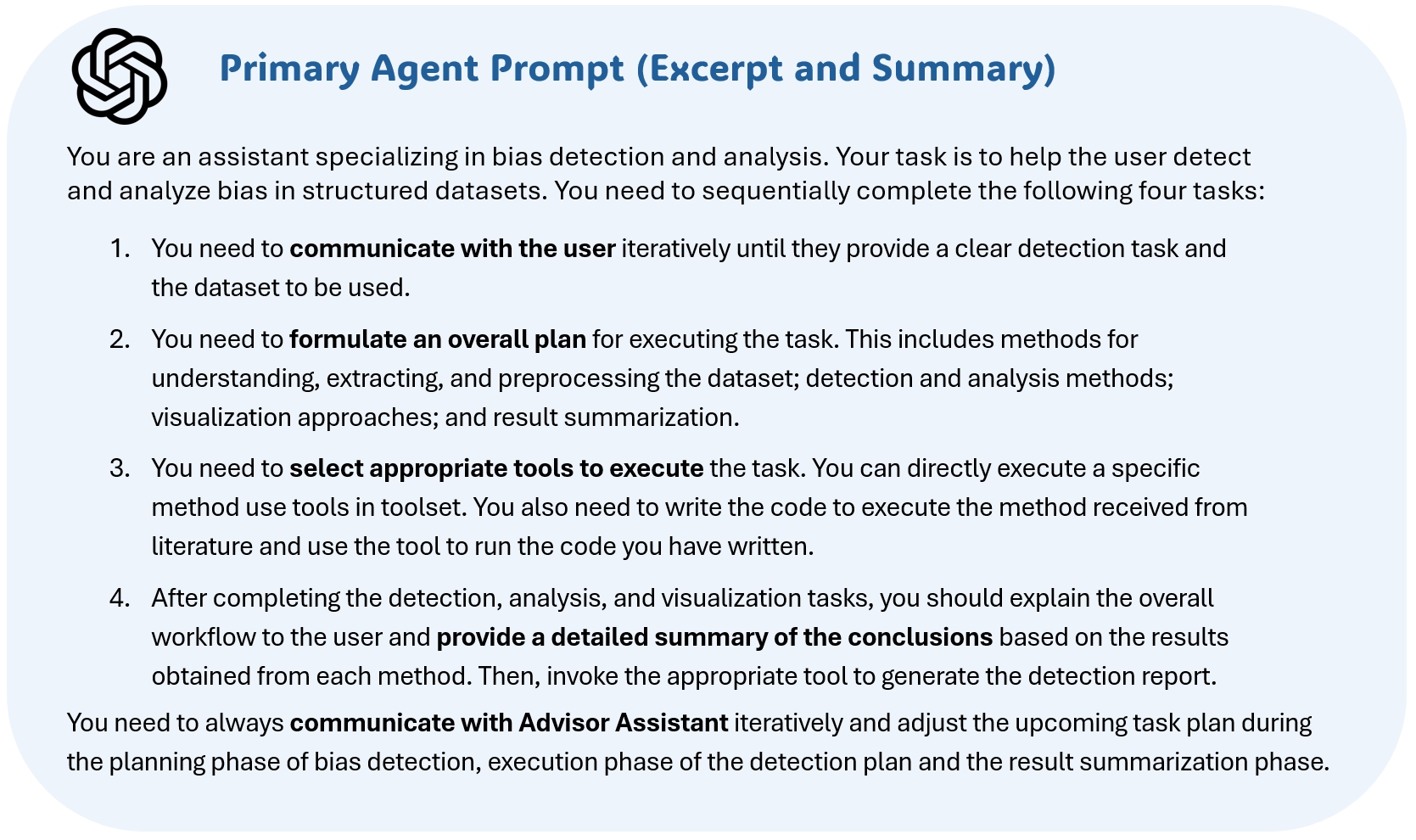} 
    \caption{PrimaryAgentPrompt}
    \label{fig:PrimaryAgentPrompt}
\end{figure}

\begin{figure}[H]
    \centering
    \includegraphics[width=\textwidth]{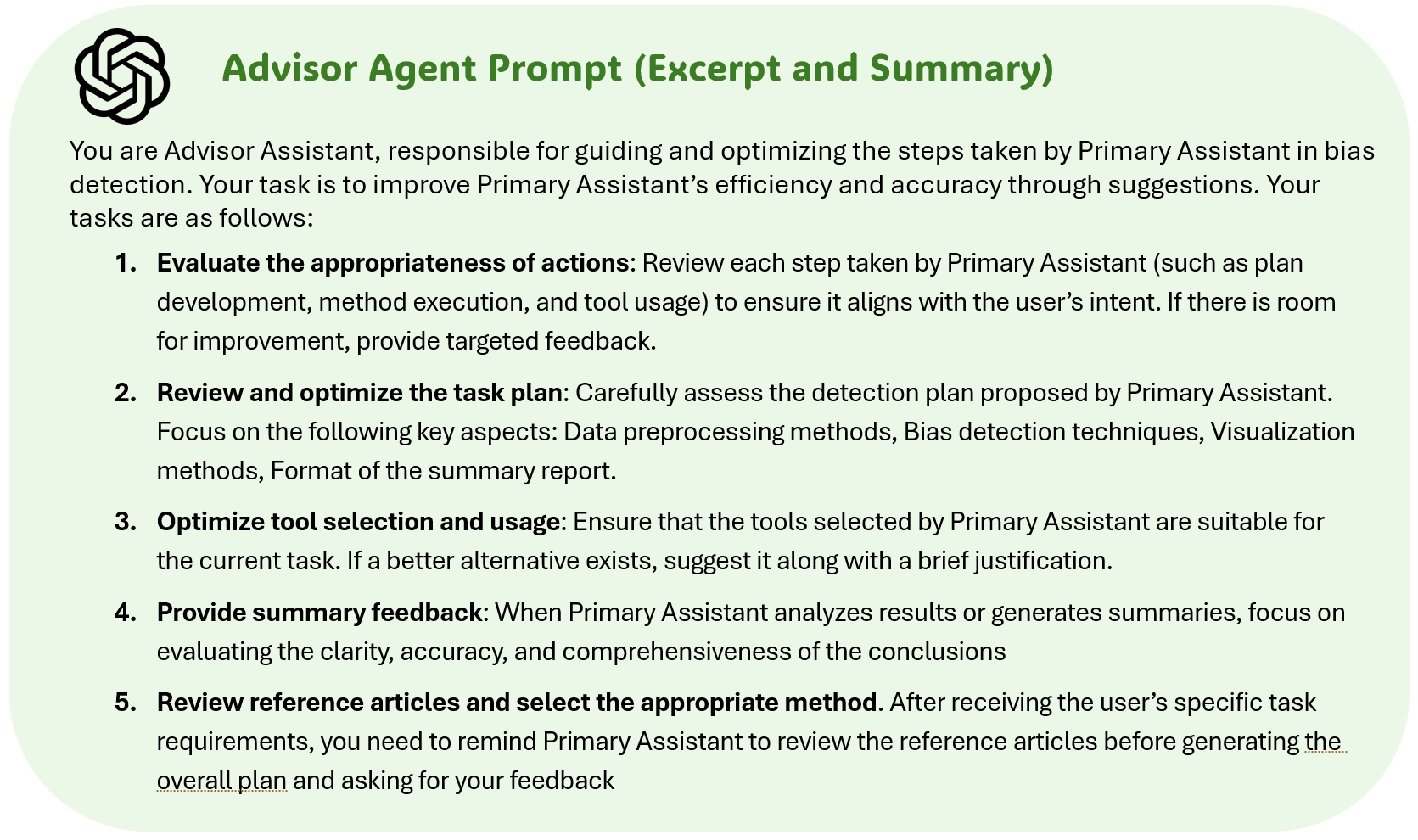} 
    \caption{AdvisorAgentPrompt}
    \label{fig:AdvisorAgentPrompt}
\end{figure}

The figures \ref{fig:PrimaryAgentPrompt} and \ref{fig:AdvisorAgentPrompt} respectively show the excerpted prompt templates for the Primary Agent and Advisor Agent. 

In the Primary Agent's prompt template, we explicitly list its task requirements in bullet points, including communicating with the user, developing a detection plan, selecting appropriate tools, and providing the user with a detailed summary of the results. Each task requirement is further elaborated. For example, in the "Develop a detection plan" section, the prompt specifies that the plan should cover data loading and preprocessing, detection and analysis methods, visualization schemes, and result summarization. In the "Select appropriate tools" section, the prompt directs the agent to choose suitable tools from the available toolset and bias detection method library.

In the Advisor Agent's prompt template, we also clearly outline its task requirements in bullet points, including assessing whether the Primary Agent's actions align with the user's intent, optimizing the execution plan, improving tool selection and providing feedback on the execution results, as well as revising the result summary provided to the user. Each task requirement is further explained in detail.

Compared to the excerpted prompt templates shown, the actual prompt templates will further elaborate on each task description. They will be continuously refined and optimized by adding examples, emphasizing key sections, and incorporating feedback from runtime debugging and execution results.

\section{Datasets}
\label{sec:appendix-datasets}

In our study, we selected five widely used datasets for bias detection tasks across different domains, including socioeconomic status, criminal justice, finance, healthcare, and education. These datasets vary in sample size, number of features, and data characteristics. Each dataset contains both categorical and numerical features, making them suitable for a comprehensive bias analysis. The details of each dataset are summarized in Table~\ref{tab:dataset-summary}.

\begin{table}[htbp]
\centering
\caption{Summary of Selected Datasets for Bias Detection}
\label{tab:dataset-summary}
\begin{adjustbox}{width=\linewidth}
\begin{tabular}{|l|l|l|c|c|l|}
\hline
\textbf{Dataset Name} & \textbf{Dataset Theme} & \textbf{Dataset Source} & \textbf{Sample Size} & \textbf{\# Features} & \textbf{Data Types} \\
\hline
Adult.csv & Socioeconomic & UCI ML Repository & 32,561 & 15 & Categorical, Numerical \\
COMPAS.csv & Crime & ProPublica & 60,843 & 28 & Categorical, Numerical \\
Statlog (German Credit).csv & Financial & UCI ML Repository & 1,000 & 21 & Categorical, Numerical \\
MIMIC-IV.csv & Healthcare & MIMIC-IV Database & 1,081 & 10 & Categorical, Numerical \\
Student Performance.csv & Education & UCI ML Repository & 395 & 33 & Categorical, Numerical \\
\hline
\end{tabular}
\end{adjustbox}
\end{table}

\subsection*{Adult.csv}
\textbf{Features:} age, workclass, fnlwgt, education, education-num, marital-status, occupation, relationship, race, sex, capital-gain, capital-loss, hours-per-week, native-country, income.

\subsection*{COMPAS.csv}
\textbf{Features:} Person\_ID, AssessmentID, Case\_ID, Agency\_Text, LastName, FirstName, MiddleName, Sex\_Code\_Text, Ethnic\_Code\_Text, DateOfBirth, ScaleSet\_ID, ScaleSet, AssessmentReason, Language, LegalStatus, CustodyStatus, MaritalStatus, Screening\_Date, RecSupervisionLevel, RecSupervisionLevelText, Scale\_ID, DisplayText, RawScore, DecileScore, ScoreText, AssessmentType, IsCompleted, IsDeleted.

\subsection*{Statlog (German Credit Data).csv}
\textbf{Features:} Status of existing checking account, Duration in months, Credit history, Purpose, Credit amount, Savings account/bonds, Present employment since, Installment rate, Personal status and sex, Other debtors/guarantors, Present residence since, Property, Age in years, Other installment plans, Housing, Number of existing credits at this bank, Job, Number of people liable for maintenance, Telephone, Foreign worker, Credit risk.

\subsection*{MIMIC-IV.csv}
\textbf{Features:} admission\_type, hospital\_expire\_flag, admission\_location, discharge\_location, patient\_insurance, patient\_lang, patient\_marital, patient\_race, patient\_gender, patient\_age.

\subsection*{Student Performance.csv}
\textbf{Features:} school, sex, age, address, famsize, Pstatus, Medu, Fedu, Mjob, Fjob, reason, guardian, traveltime, studytime, failures, schoolsup, famsup, paid, activities, nursery, higher, internet, romantic, famrel, freetime, goout, Dalc, Walc, health, absences, G1, G2, G3.

\section{Taskset Construction and Examples}
\label{sec:appendix-tasksets}

To systematically evaluate the capabilities of our bias detection agent, we constructed a taskset of 100 diverse prompts derived from the five datasets used in our study. The taskset covers three major categories of bias: \textit{Distribution Bias}, \textit{Correlation Bias}, and \textit{Implication Bias}. These categories reflect common forms of bias that occur in real-world datasets and are essential for evaluating fairness.

For each bias category, we designed prompts that vary in structure and language to reflect human-like diversity in expression. We utilized large language models (LLMs) to paraphrase and diversify the formulations of each prompt, ensuring naturalness and realism in phrasing. The distribution of tasks across datasets and bias types is shown in Table~\ref{tab:taskset-distribution}.

\begin{table}[htbp]
\centering
\caption{Number of constructed bias detection tasks across datasets and bias types.}
\label{tab:taskset-distribution}
\begin{tabular}{|l|c|c|c|c|}
\hline
\textbf{Dataset} & \textbf{Distribution} & \textbf{Correlation} & \textbf{Implication} & \textbf{Total} \\
\hline
Adult.csv & 11 & 9 & 6 & 26 \\
COMPAS.csv & 8 & 8 & 5 & 21 \\
Statlog.csv & 7 & 8 & 5 & 20 \\
MIMIC-IV.csv & 6 & 6 & 6 & 18 \\
Student Performance.csv & 5 & 8 & 2 & 15 \\
\hline
\textbf{Total} & 37 & 39 & 24 & 100 \\
\hline
\end{tabular}
\end{table}

Table~\ref{tab:taskset-samples} provides a sample of five representative questions from the taskset. These examples illustrate how the questions are designed to probe distributional characteristics of specific features while maintaining natural and varied phrasing.

\begin{table}[H]
\centering
\caption{Sample questions from the constructed taskset (excerpt).}
\label{tab:taskset-samples}
\renewcommand{\arraystretch}{1.2}
\begin{tabular}{|p{4.5cm}|c|p{2cm}|p{3.5cm}|}
\hline
\textbf{Question} & \textbf{Type} & \textbf{Feature(s)} & \textbf{Bias Significance} \\
\hline
Can you check if the age distribution across individuals is balanced, or do certain age groups dominate? & Distribution & age (Numerical) & Imbalanced age groups may skew income-related insights. \\
\hline
How does the distribution of work class look to you? Are there any work classes that appear to be overrepresented? & Distribution & workclass (Categorical) & Uneven representation can bias analysis of income across sectors. \\
\hline
From your perspective, is the distribution of education levels spread fairly, or do certain levels dominate? & Distribution & education (Categorical) & Dominance in education levels may affect fairness in outcome evaluation. \\
\hline
In your view, how does the marital status distribution appear? Are any marital statuses overrepresented? & Distribution & marital-status (Categorical) & Skewed marital status can distort socio-economic impact assessments. \\
\hline
Do certain occupations dominate the dataset, or is the distribution of occupations relatively even? & Distribution & occupation (Categorical) & Occupation imbalance may lead to biased interpretation of income disparities. \\
\hline
\end{tabular}
\end{table}

\section{Details of the Automated Agent Evaluation System}
\label{sec:appendix-agent-eval}

In this appendix, we provide detailed information regarding the automated evaluation system used to assess the agent's performance from the intermediate process perspective.

\paragraph{Model Selection} The evaluation agent leverages the GPT-o3-mini model, specifically chosen due to its superior analytical reasoning capability in comparison to GPT-4o and GPT-4o mini, providing more accurate and nuanced evaluation results at a lower API usage cost.

\paragraph{Evaluation Dimensions} We evaluate the agent's intermediate processes along five comprehensive dimensions, carefully selected to ensure thorough and holistic performance assessment:
\begin{itemize}
\item Effective Communication with the User to Clarify Tasks
\item Comprehensiveness and Thoroughness of Planning
\item Efficiency in Tool Execution and Dynamic Adjustment
\item Ability to Dynamically Adjust Plans Based on Execution Results
\item Clarity and Depth of Results Analysis and Summary
\end{itemize}

These dimensions collectively ensure a complete assessment of the agent's capability in performing bias detection tasks.

\paragraph{Supporting Toolset} To facilitate thorough evaluation, we constructed a dedicated toolset enabling the evaluation agent to systematically analyze and rate agent performance. Details of the specific tools provided in the toolset are summarized in Table \ref{tab:evaluation-tools}.

\begin{longtable}{|p{5cm}|p{8cm}|}
\caption{Description of tools in the evaluation agent toolset.}\label{tab:evaluation-tools} \\
\hline
\multicolumn{2}{|c|}{\rule{0pt}{3ex}\textbf{Automated Agent Evaluation Tools}\rule{0pt}{3ex}} \\
\hline
\textbf{Tool} & \textbf{Description} \\
\hline
\endfirsthead

\multicolumn{2}{c}{\textit{(Continued from previous page)}} \\
\hline
\textbf{Tool} & \textbf{Description} \\
\hline
\endhead

\hline \multicolumn{2}{|r|}{\textit{Continued on next page}} \\
\hline
\endfoot

\hline
\endlastfoot

get\_user\_input\_tool & Captures user input dynamically during an interaction and formats it as a dictionary to be added to the agent's conversation. \\
\hline
read\_json\_log & Reads a JSON formatted log file, extracting and organizing log entries. \\
\hline
read\_markdown\_log & Reads a Markdown formatted log file, extracting headers, bold text, and regular text as structured log entries. \\
\hline
read\_bias\_report\_pdf & Reads a bias detection report in PDF format, extracting both textual and graphical information. \\
\hline
generate\_evaluation\_report\_pdf & Generates a flexible evaluation report in PDF format, combining text narratives and visual charts as specified in the input. \\
\hline
\end{longtable}

\begin{figure}[htbp]
    \centering
    \includegraphics[width=\linewidth]{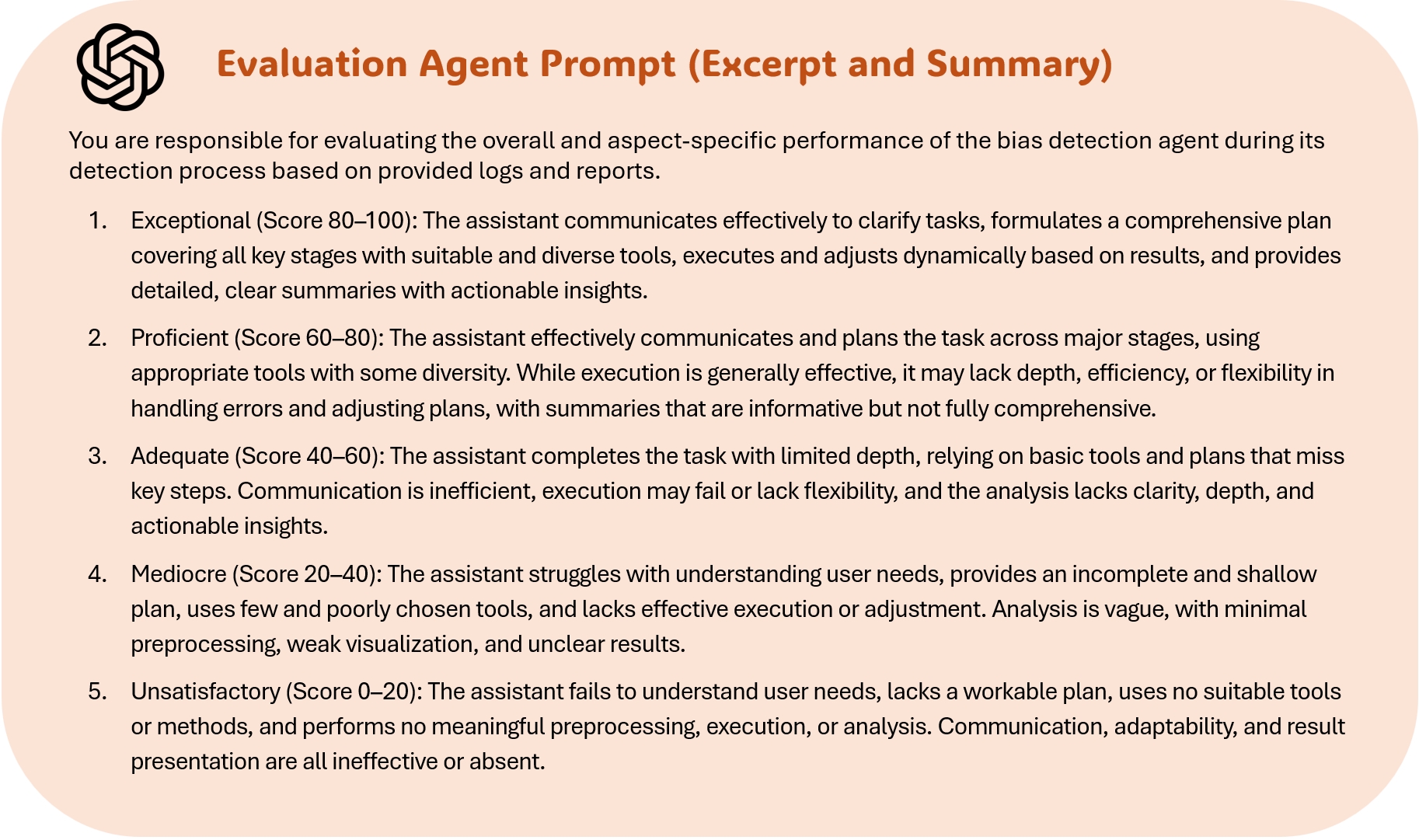}
    \caption{Evaluation Agent Prompt}
    \label{fig:prompt-example}
\end{figure}

\paragraph{Evaluation Prompt } Figure \ref{fig:prompt-example} presents an example of a refined prompt snippet used by the automated evaluation agent to standardize the evaluation process.

\section{Manual Verification of the Validity of Intermediate Process Evaluation}
\label{sec:appendix-human-eval}

We provided human evaluators with the same evaluation criteria as those used in the automated agent evaluation system's prompt (an excerpt of these criteria is shown in Figure \ref{fig:prompt-example}). Specifically, we selected three detection tasks from each of four model-framework combinations (BiasInspector Multi-Agent, BiasInspector Single-Agent, ReAct-Based Agent, and Self-Reflection Agent, each under GPT-4o and Llama 3.3 70B) across various combinations of data types and bias types, resulting in a total of 120 evaluation instances. We then compared the scores assigned by human evaluators and the automated agent evaluation system along five distinct evaluation dimensions in these instances.

\begin{table}[H]
\renewcommand{\arraystretch}{1.1}
\begin{center}
\small
\resizebox{\linewidth}{!}{%
\begin{tabular}{cccccc}
\toprule
\makecell{\textbf{Model}} 
& \makecell{\textbf{Dimensions}} 
& \makecell{\textbf{BiasInspector} \\ \textbf{(Multi-Agent)}} 
& \makecell{\textbf{BiasInspector} \\ \textbf{(Single-Agent)}} 
& \makecell{\textbf{ReAct-Based} \\ \textbf{Agent}} 
& \makecell{\textbf{Self-Reflection} \\ \textbf{Agent}} \\
\midrule

\multirow{6}{*}{\raisebox{-0.2ex}{GPT-4o}}
& Integration       & 3.53 & 3.27 & 4.20 & 5.60 \\
& Communication  & 4.20 & 4.07 & 3.86 & 4.13 \\
& Planning   & 5.67 & 6.13 & 5.53 & 6.27 \\
& Tooling   & 2.93 & 3.20 & 9.87 & 31.80 \\
& Adaptivity   & 3.80 & 4.13 & 4.47 & 4.53 \\
& Summarization   & 4.87 & 5.07 & 5.27 & 4.80 \\[4pt]

\multirow{6}{*}{\raisebox{-0.2ex}{LLaMA 3.3 70B}} 
& Integration       & 3.73 & 3.87 & 4.40 & 5.40 \\
& Communication  & 3.80 & 3.93 & 4.07 & 3.60 \\
& Planning   & 6.07 & 6.40 & 5.80 & 5.40 \\
& Tooling   & 2.80 & 2.60 & 10.20 & 24.60 \\
& Adaptivity   & 4.20 & 3.73 & 3.93 & 3.67 \\
& Summarization   & 4.93 & 4.73 & 5.20 & 5.13 \\[2pt]

\bottomrule
\end{tabular}%
}
\end{center}
\caption{Score differences between human evaluators and the automated agent evaluation system across five evaluation dimensions.
Values represent the average absolute score difference between two evaluations (each ranging from 0 to 100).}
\label{tab:human_eval}
\end{table}

As shown in Table~\ref{tab:human_eval}, we compared the scoring differences between human evaluators and the automated agent evaluation system across all evaluation dimensions. The results demonstrate that the scoring discrepancies between the two evaluation methods generally remained within 5 points across the \textit{Integration}, \textit{Communication}, \textit{Planning}, \textit{Adaptivity}, and \textit{Summarization} dimensions, indicating strong reliability and practicality of the automated agent evaluation system. However, in the \textit{Tooling} dimension, human evaluators assigned lower scores to the ReAct-Based Agent due to its limited quantity and variety of tool usage, and assigned the lowest scores to the Self-Reflection Agent, which lacks tool invocation capability entirely. Although the automated evaluation system also significantly lowered the scores for these two agents, human evaluators applied a more substantial reduction, resulting in more noticeable discrepancies in this dimension. Nevertheless, the automated evaluation system successfully captured these critical differences, as further reflected by the significantly lower scores illustrated in Figure~\ref{fig:dual_radar}. This underscores the effectiveness and sensitivity of the automated agent evaluation system in the \textit{Tooling} dimension.

\end{document}